\documentclass[journal,twoside,web]{ieeecolor}
\usepackage{jsen}
\usepackage{cite}
\usepackage{amsmath,amssymb,amsfonts}
\usepackage{algorithmic}
\usepackage{graphicx}
\usepackage{textcomp}
\usepackage{wrapfig}
\usepackage{mathrsfs}
\usepackage[ruled,vlined]{algorithm2e}
\usepackage{graphicx}
\usepackage{booktabs}
\usepackage{blkarray}
\usepackage{ulem}

\def\BibTeX{{\rm B\kern-.05em{\sc i\kern-.025em b}\kern-.08em
    T\kern-.1667em\lower.7ex\hbox{E}\kern-.125emX}}
\definecolor{abstractbg}{rgb}{0.89804,0.94510,0.83137}
\begin{document}

\title{An Efficient Calibration Method for Triaxial Gyroscope}
\author{Li Wang, Tao Zhang, Lin Ye, Jiao Jiao Li,
Steven Su,~\IEEEmembership{Senior Member,~IEEE}
\thanks{L. Wang, J. J. Li and S. W. Su is with Faculty of Engineering and IT, University of Technology Sydney, Sydney, NSW 2007, Australia (e-mail: li.wang-9@student.uts.edu.au, jiaojiao.li@uts.edu.au; steven.su@uts.edu.au).}
\thanks{T. Zhang is with Multi-Scale Medical Robotics Center and Chow Yuk Ho Technology Centre for Innovative Medicine, The Chinese University of Hong Kong, Shatin, N.T., Hong Kong. (e-mail: Tao.Zhang-4@alumni.uts.edu.au).}
\thanks{L. Ye is with Enjoymove Technology Co., Ltd, Shanghai, China. (e-mail: lin.ye@enjoymove.cn).}
}

\IEEEtitleabstractindextext{%
\fcolorbox{abstractbg}{abstractbg}{%
\begin{minipage}{\textwidth}%
\begin{wrapfigure}[12]{r}{3in}%
\includegraphics[width=2.9in, height=1.4in]{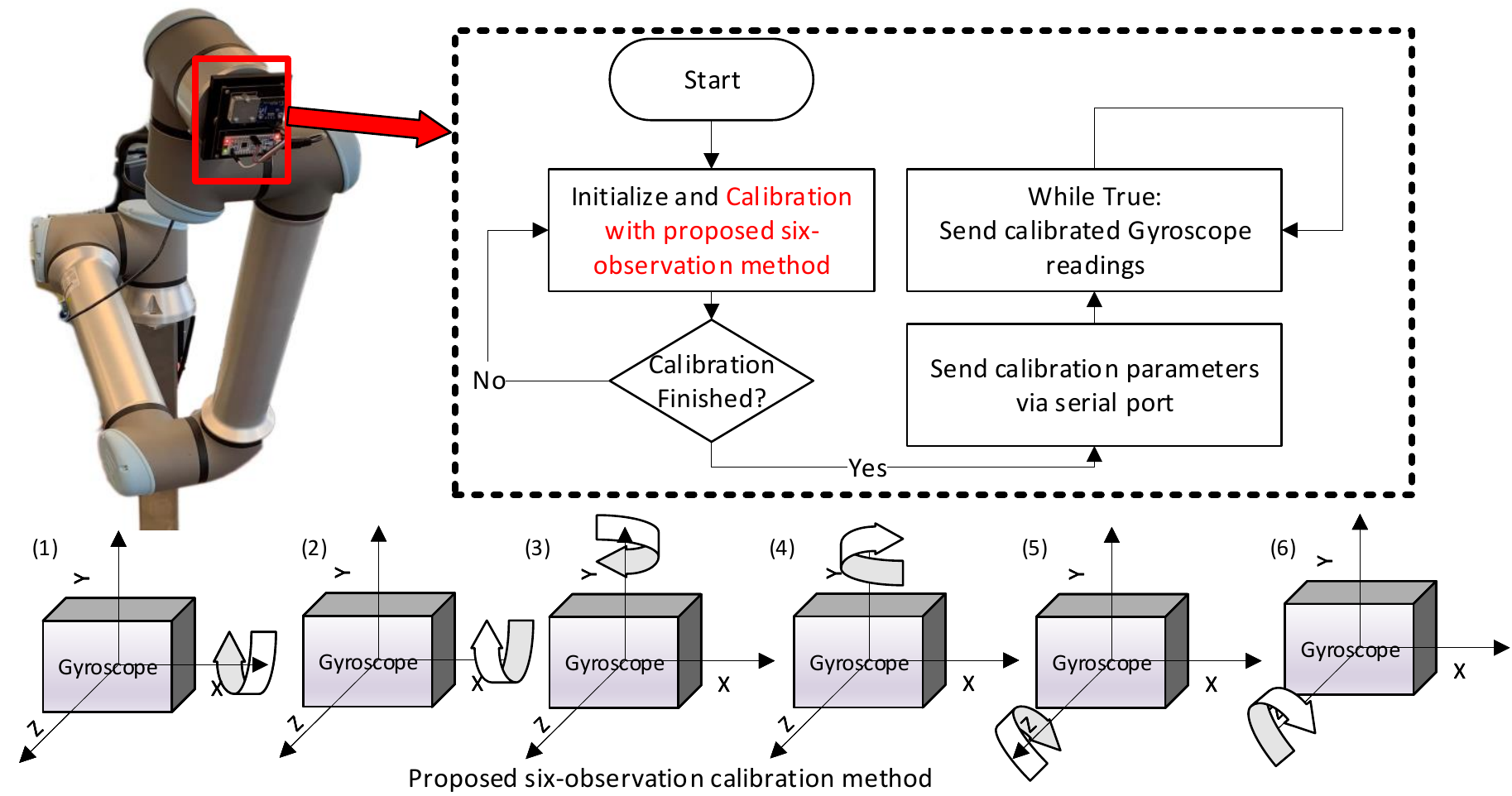}%
\end{wrapfigure}%
\begin{abstract}
This paper presents an efficient servomotor-aided calibration method for the triaxial gyroscope. The entire calibration process only requires approximately one minute, and does not require high-precision equipment. This method is based on the idea that the measurement of the gyroscope should be equal to the rotation speed of the servomotor. A six-observation experimental design is proposed to minimize the maximum variance of the estimated scale factors and biases. In addition, a fast converging recursive linear least square estimation method is presented to reduce computational complexity. The simulation results reflect the robustness of the calibration method under normal and extreme conditions. We experimentally demonstrate the feasibility of the proposed method on a robot arm, and implement the method on a microcontroller. We verify the calibration results of the proposed method by comparing with a traditional turntable approach, and the experiment indicates that the results of these two methods are comparable. By comparing the calibrated low-cost gyroscope reading with the reading from a high-precision gyroscope, we can conclude that our method significantly increases the gyroscope’s accuracy. 
\end{abstract}

\end{minipage}}}

\maketitle

\section{Introduction}
\label{sec:introduction}
\IEEEPARstart{T}{he} Micro-electro-mechanical-system (MEMS) triaxial gyroscopes are commonly used devices for measuring angular velocity in a broad range of applications, such as indoor pedestrian positioning \cite{indoor1}, health monitoring \cite{healthMonitoring}, and consumer electronic devices \cite{consumer1,consumer2}. Such low-cost gyroscopes usually do not show high precision, due to the accumulation of drift error from integration when calculating the attitude \cite{errorAccumulate}. In addition, the low repeatability and instability of the gyroscope cause changes in the scale factor and biases on every boot or under different environmental conditions, such as temperature variation \cite{errorTemperature}. Therefore, the gyroscope needs to be calibrated before each use or when the environmental conditions change. Simple and efficient calibration methods are required for frequent calibration to be practical.

The issue of gyroscope calibration has received considerable attention. The ordinary triaxial gyroscope calibration method involves rotating the gyroscope at known angular velocity \cite{conventionalmethod}. This approach can achieve high calibration accuracy, but the complex calibration procedure and requirement for expensive equipment make it unsuitable for use outside the laboratory. Gyroscope calibration methods that do not require precise rotation velocity measurements are presented in \cite{camera,mag,acc1,acc2}. In \cite{camera}, a camera is employed to provide position and orientation information for gyroscope calibration. This technique first requires alignment of the body frame and the image frame, and calibration of the camera to achieve high accuracy. As images are involved, the computational complexity is significantly increased. 
{In \cite{mag}, a magnetometer-aided calibration method is investigated. The gyroscope calibration reference is provided by a homogeneous magnetic field. However, a weak magnetic field (e.g., the local magnetic field) can be easily disturbed by external alternating magnetic fields such as power lines. Therefore, this method may not be suitable for certain in-field applications.} 
In \cite{acc1,acc2}, accelerometer-aided gyroscope calibration methods are proposed. The accelerometer is first calibrated by gravity using the multi-position method, which then provides the rotation speed of the gyroscope. The entire calibration procedure takes more than 10 minutes, which is cumbersome in practical operation. Besides, the error caused by the accelerometer may be superimposed on the gyroscope parameters. Therefore, there is a great need to find a fast and simple in-field calibration method for gyroscopes. 
This paper proposes a servomotor-aided gyroscope calibration method, which does not require high precision equipment and is easy to implement outside the laboratory.

To meet the requirements of frequent calibration, improving the calibration efficiency is of primary importance. Most previous studies \cite{conventionalmethod,camera,mag,acc1,acc2,triaxialMathematicalModel} paid little attention to investigating the selection of optimal experimental design. Recently, a six-position accelerometer calibration experimental design (DoE) was proposed \cite{ye2017efficient}. The purpose of DoE is to obtain sufficient information for calibrating the accelerometer using a minimum number of experiments. To the best of our knowledge, this is the first study to apply such DoE to gyroscope calibration. In this study, we propose a G-optimal DoE for gyroscope calibration, minimizing the maximum estimation variance over the entire measurement range.

For solving the regression problem, \cite{won2009triaxial,acc2,qureshi2017algorithm} applied nonlinear estimation methods, such as the Nelder–Mead method\cite{NelderMead} and Levenberg-Marquardt algorithm \cite{LevenbergMarquardt}. These methods typically require sizeable computational power, which are difficult to apply for gyroscope calibration since a gyroscope is often part of an embedded system with limited resources and battery life. Therefore, we propose a fast converging recursive linear least square estimation for the six-parameter gyroscope calibration model.

We summarise the contributions of this paper as follows. First, we propose an efficient servomotor-aided gyroscope calibration method, which does not require the use of high precision equipment during calibration. Second, we propose a six-point G-optimal experiment for gyroscope calibration. The proposed DoE can significantly shorten the calibration time and has been empirically validated. Last, we implement a fast converging recursive linear least square estimation method to reduce the computational complexity, which makes the calibration process more adaptable for an embedded environment.

This paper is organized as follows. In section \ref{CalibrationMethodology}, we discuss the proposed calibration method and the DoE. In Section \ref{Sec:Simulation}, we validate the approach using simulations under different conditions. In Section \ref{Sec:Experiment}, we demonstrate the implementation of the proposed method on two commonly used gyroscopes. Section \ref{Sec:Conclusion} concludes this paper.

\section{Calibration methodology} \label{CalibrationMethodology}
\subsection{Efficient Calibration Method for Triaxial Gyroscope}
Various factors contribute to error in a gyroscope. In this study, scale factors and biases are considered as error sources. Thus, a six-parameter calibration model is employed to define the unknown parameters. The relation between the actual angular velocity $\mathbf{G}_i=[g_{x,i}, g_{y,i}, g_{z,i}]^T$ and the measured angular velocity $\mathbf{M}_i=[m_{x,i}, m_{y,i}, m_{z,i}]^T$ at the $i$ th observation is described as:

\begin{equation}\label{model}
 \begin{bmatrix}
g_{x,i} \\
g_{y,i} \\
g_{z,i}
\end{bmatrix}
=
\begin{bmatrix}
k_x & 0 & 0 \\
0 & k_y & 0 \\
0 & 0 & k_z
\end{bmatrix}
\begin{pmatrix}
  \begin{bmatrix}
    m_{x,i} \\
    m_{y,i} \\
    m_{z,i}
  \end{bmatrix} + \begin{bmatrix}
                    b_x \\
                    b_y \\
                    b_z
                  \end{bmatrix}
\end{pmatrix}
\end{equation}
where $k_x,k_y,k_z$ and $b_x,b_y,b_z$ stand for the scale factor and the bias of each axis, respectively.

This method is based on the idea that the measurement of the gyroscope should be equal to the rotation speed, that is
\begin{equation}\label{sumofsqure}
  \omega_{i} = \sqrt{g_{x,i}^{2}+g_{y,i}^{2}+g_{z,i}^{2}}
\end{equation}
where $\omega_i$ is the rotation speed. 

We can expand Eq.\eqref{sumofsqure} and square both sides of the equation. Then, we have:
\begin{equation}\label{expand}
  \begin{split}
     \omega_i^2 & =k_x^2 m_{x,i}^2 + k_y^2 m_{y,i}^2 +k_z^2 m_{z,i}^2 \\
       & +2k_x^2 b_x m_{x,i}+2k_y^2 b_y m_{y,i}+2k_z^2 b_z m_{z,i}\\
       & +\sum_{j=x,y,z}k_j^2 b_j^2+\epsilon_i .
  \end{split}
\end{equation}
The error term $\epsilon_i$ is a combination of a Gaussian and a noncentral Chi-squared noise. Similar to \cite{noiseprove}, the Chi-squared noise term can be ignored when the rotating speed is high. Thus, in this study, we consider the $\epsilon_i$ as a Gaussian noise.
If we let
\begin{flalign*}
      \left\{
        \begin{array}{l}
            \beta_0=\sum_{j=x,y,z} k_j^2 b_j^2\\
            \beta_1=k_x^2\\
            \beta_2=k_y^2\\
            \beta_3=k_z^2\\
            \beta_4=2 k_x^2 b_x\\
            \beta_5=2 k_y^2 b_y\\
            \beta_6=2 k_z^2 b_z
        \end{array}
    \right. &       \left\{
        \begin{array}{l}
            x_{1,i}=m_{x,i}^2\\
            x_{2,i}=m_{y,i}^2\\
            x_{3,i}=m_{z,i}^2\\
            x_{4,i}=m_{x,i}\\
            x_{5,i}=m_{y,i}\\
            x_{6,i}=m_{z,i}\\
        \end{array}
    \right.  &
            y_i=\omega_i^2,
\end{flalign*}
then the gyroscope calibration problem becomes
\begin{equation}\label{linearregression}
\begin{split}
        y_{i}=&\beta_0+\beta_1 x_{1,i}+\beta_2 x_{2,i}+\beta_3 x_{3,i}\\
        &+\beta_4 x_{4,i} +\beta_5 x_{5,i}+\beta_6 x_{6,i}+\epsilon_i,
\end{split}
\end{equation}
which is a linear regression problem. The cost function of this problem is defined as:
\begin{equation}\label{costfunc}
J=\sum_{i=1}^{n}(||y_i-y_{actual,i}||) ,
\end{equation}
where $y_{actual}$ is the squared rotation speed provided by the servomotor. However, there is no close form solution for this problem as the parameter $\beta_0$ can be represented by the remaining six parameters $\beta_0=\frac{\beta_4^2}{4\beta_1}+\frac{\beta_5^2}{4\beta_2}+\frac{\beta_6^2}{4\beta_3}$. The representation introduces the nonlinearity to Eq.\eqref{linearregression}. Thus, the question can be solved using a nonlinear regression technique such as the Levenberg-Marquardt algorithm \cite{LevenbergMarquardt}. As mentioned above, a gyroscope is often part of an embedded system with limited resources. To reduce the computational complexity, a novel iterative least square method \cite{ye2017efficient} is employed to estimate the parameters.

We can reform Eq.\eqref{linearregression} in matrix form as:
\begin{equation}\label{matrix}
  Y=X\boldsymbol{\beta}+\boldsymbol{\beta_0}+\boldsymbol{\epsilon} .
\end{equation}
The observation matrix $X\in\mathbb{R}^{6\times6}$ consists of the measured angular velocity:
\begin{equation}\label{observationMatrix}
X=
  \begin{bmatrix}
  m_{x,1}& m_{y,1}& m_{z,1}&m_{x,1}^2& m_{y,1}^2& m_{z,1}^2\\
m_{x,2}& m_{y,2}& m_{z,2}&m_{x,2}^2& m_{y,2}^2& m_{z,2}^2\\
m_{x,3}& m_{y,3}& m_{z,3}&m_{x,3}^2& m_{y,3}^2& m_{z,3}^2\\
m_{x,4}& m_{y,4}& m_{z,4}&m_{x,4}^2& m_{y,4}^2& m_{z,4}^2\\
m_{x,5}& m_{y,5}& m_{z,5}&m_{x,5}^2& m_{y,5}^2& m_{z,5}^2\\
m_{x,6}& m_{y,6}& m_{z,6}&m_{x,6}^2& m_{y,6}^2& m_{z,6}^2
  \end{bmatrix}.
\end{equation}
The response matrix $Y=[y_{actual,1},y_{actual,2},\ldots,y_{actual,6}]^T$, parameters $\boldsymbol{\beta}=[\beta_1,\beta_2,\ldots,\beta_6]^T$, bias term $\boldsymbol{\beta_0}\in\mathbb{R}^{1\times6}=[\beta_0,\beta_0,\ldots,\beta_0]^T$, and noise term $\boldsymbol{\epsilon}=[\epsilon_1,\epsilon_2,\ldots,\epsilon_6]^T$.

For solving Eq.\eqref{matrix}, the fast converging iterative least square method is summarized in Algorithm \ref{algorithm}. The convergent condition is given as $0\leq\beta_0<0.5$ \cite{ye2017efficient}. For a low cost MEMS gyroscope, the scale factor is usually within the range of $[0.8, 1.2]$. The bias term is usually between $\pm 0.1 rad/s$. Recall that $\beta_0=\sum_{j=x,y,z} k_j^2 b_j^2$. Obviously, the convergence condition is met.
\begin{algorithm}\label{algorithm}
\SetAlgoLined
\KwResult{Estimated scale factors $[k_x, k_y, k_z]$ and bias $[b_x, b_y, b_z]$.}
 Set initial value $\boldsymbol{\beta_0}^{(0)}=[0,0,0,0,0,0]$;\\
 Calculate initial estimation $\boldsymbol{\beta}^{(1)}=(X^T X)^{-1}X^T (Y-\boldsymbol{\beta_0}^{(0)})$;\\
 \While{$\sum_{j=1}^{6}||{\beta_j}^{(n+1)}-{\beta_j}^{(n)}||>10^{-6}$}{
  Update $\boldsymbol{\beta_0}$ at $n$th iteration as follows:
  ${{\gamma}}^{(n)}=\frac{{\beta_4^2}^{(n)}}{4{\beta_1}^{(n)}}+\frac{{\beta_5^2}^{(n)}}{4{\beta_2}^{(n)}}+\frac{{\beta_6^2}^{(n)}}{4{\beta_3}^{(n)}}$;\\
  $\boldsymbol{\beta_0}^{(n)}=[{{\gamma}^{(n)},{\gamma}^{(n)},{\gamma}^{(n)},{\gamma}^{(n)},{\gamma}^{(n)},{\gamma}^{(n)}}]$ ;\\
  Update $\boldsymbol{\beta}$ at $n$th iteration as follows:\\
  $\boldsymbol{\beta}^{(n+1)}=(X^T X)^{-1}X^T (Y-\boldsymbol{\beta_0}^{(n)})$.
 }
 \KwRet Scale factors and bias terms:
\begin{flalign*}
      \left\{
        \begin{array}{l}
            k_x=\sqrt{\beta_1}\\
            k_y=\sqrt{\beta_2}\\
            k_z=\sqrt{\beta_3}
        \end{array}
    \right. &       \left\{
        \begin{array}{l}
            b_x=\frac{\beta_4}{2\beta_1}\\
            b_y=\frac{\beta_5}{2\beta_2}\\
            b_z=\frac{\beta_6}{2\beta_3}
        \end{array}
    \right.
\end{flalign*}
 \caption{Iterative least square method}
\end{algorithm}

\begin{figure}
	\centering
		\includegraphics[scale=.5]{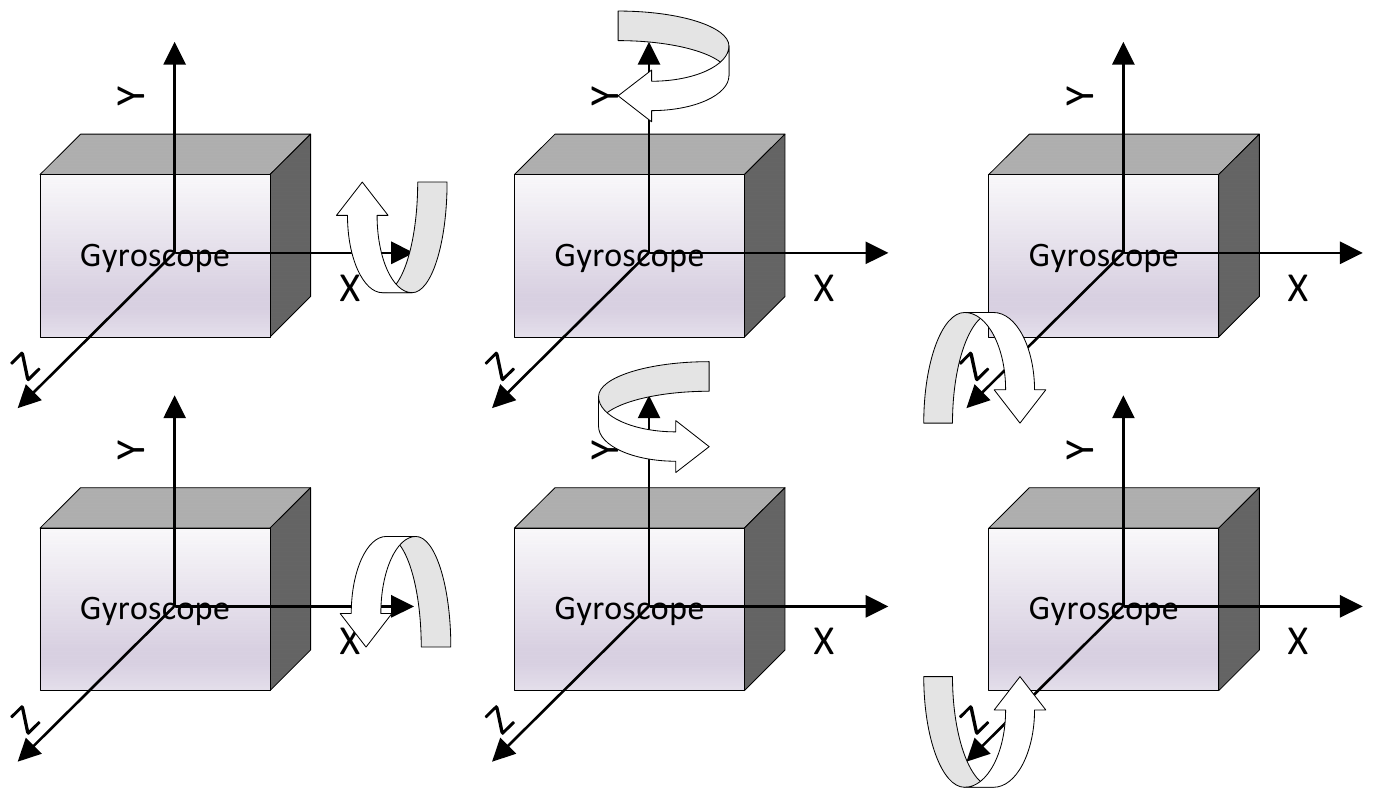}
	\caption{{Six-observations rotation protocol for gyroscope calibration. The gyroscope is rotated at constant speed clockwise and counterclockwise along the x,y,z axis.}}
	\label{rotationmethod}
\end{figure}

\subsection{G-Optimal Experimental Design}
Unlike an accelerometer, it is difficult for a low-cost MEMS gyroscope to perform autocalibration using the Earth's rotation. The Earth rotates at a moderate angular velocity of $7.29 \times {10}^{-5} rad/s$, which is much lower than the bias term of the gyroscope. Hence, this study employs {a servomotor as an external device. Considering its working principle, a servomotor may have vibrations during operation, but the time taken per revolution is still highly accurate. Owing to the linearity of Eq.\eqref{linearregression}, we can take the average of both sides of the equation during each revolution and consider it as one observation: }

\begin{equation}\label{avglinearregression}
\begin{split}
        {\frac{1}{N}\sum_{j=1}^N y_{i,j}=\frac{1}{N}\sum_{j=1}^N(}&{\beta_0+\beta_1 x_{1,i,j}+\beta_2  x_{2,i,j}+\beta_3  x_{3,i,j}}\\
        &{+\beta_4 x_{4,i,j} +\beta_5  x_{5,i,j}+\beta_6  x_{6,i,j}+\epsilon_i).}
\end{split}
\end{equation}
{In this case, we can minimize the influences of vibrations and random noise to the estimated parameter.}

The linear regression problem includes the estimation of six parameters. Thus, at least six observations are required \cite{expnumber}. To minimize the maximum variance of the estimated parameters, we introduce a G-optimal design of a second-order three variables model Eq. \eqref{linearregression} for gyroscope calibration experiments. As the measurement is limited by $m_{x,i}^2+ m_{y,i}^2+m_{z,i}^2\approx \omega_i^2$, the design region is spherical. For a six-observations experimental scheme, the G-optimal design matrix can be expressed as:
\[D=
\begin{blockarray}{cccc}
& m_{x,i} & m_{y,i} & m_{z,i} \\
\begin{block}{c[ccc]}
   (1) & 1 & 0 & 0  \\
   (2) & -1 & 0 & 0  \\
    (3)& 0 & 1 & 0  \\
    (4)& 0 & -1 & 0  \\
    (5)& 0 & 0 & 1 \\
    (6)& 0 & 0 & -1\\
\end{block}
\end{blockarray}
 \]
 
Accordingly, the rotation method of the gyroscope is shown in Fig.\ref{rotationmethod}. We rotate the gyroscope 360 degrees clockwise and counterclockwise along the x, y, z axis at the speed of $\omega$, respectively. We average the data during each rotation and construct the 6-by-6 observation matrix according to Eq.\eqref{observationMatrix}. It is worth noting that no high-precision device is used to eliminate the alignment error. Once this six-observation matrix is constructed, the scale factors and bias terms can be calculated using Algorithm \ref{algorithm}.
\section{Simulation} \label{Sec:Simulation}
\begin{figure}
	\centering
		\includegraphics[scale=.5]{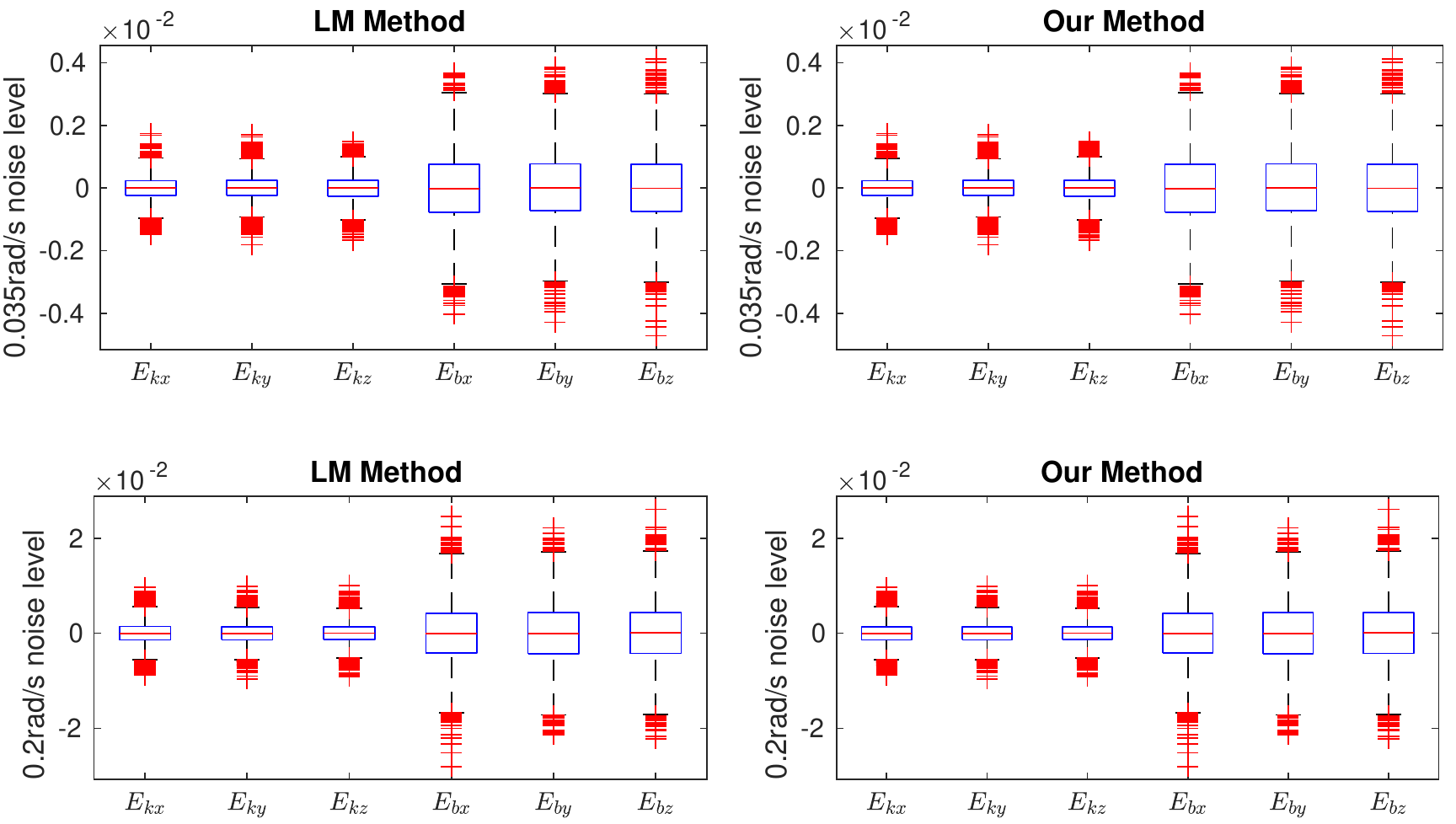}
	\caption{{The simulation results of estimation error between estimated and actual parameters under normal conditions at different noise level using different method. Top: 0.035 rad/s noise levels. Bottom: 0.2 rad/s noise level. Left: Levenberg–Marquardt (LM) method. Right: Our proposed method.}}
	\label{boxplot}
\end{figure}
With the intention of validating the feasibility of the proposed calibration method under different weights of scale factors, biases, rotation speed, orientation misalignment, and noise level, we first examined the proposed method using simulations. During each simulation, we generated a set of parameters under certain conditions, and these parameters were considered as the ground truth. Based on the actual value, the measurements of six observations were generated according to the experimental protocol in Fig.\ref{rotationmethod}. Then, the proposed method was employed to calculate the scales and biases based on the generated measurements, and the estimated parameters were stored.

\subsection{Simulation Under Normal Conditions}
\begin{figure}
	\centering
		\includegraphics[width=0.48\textwidth]{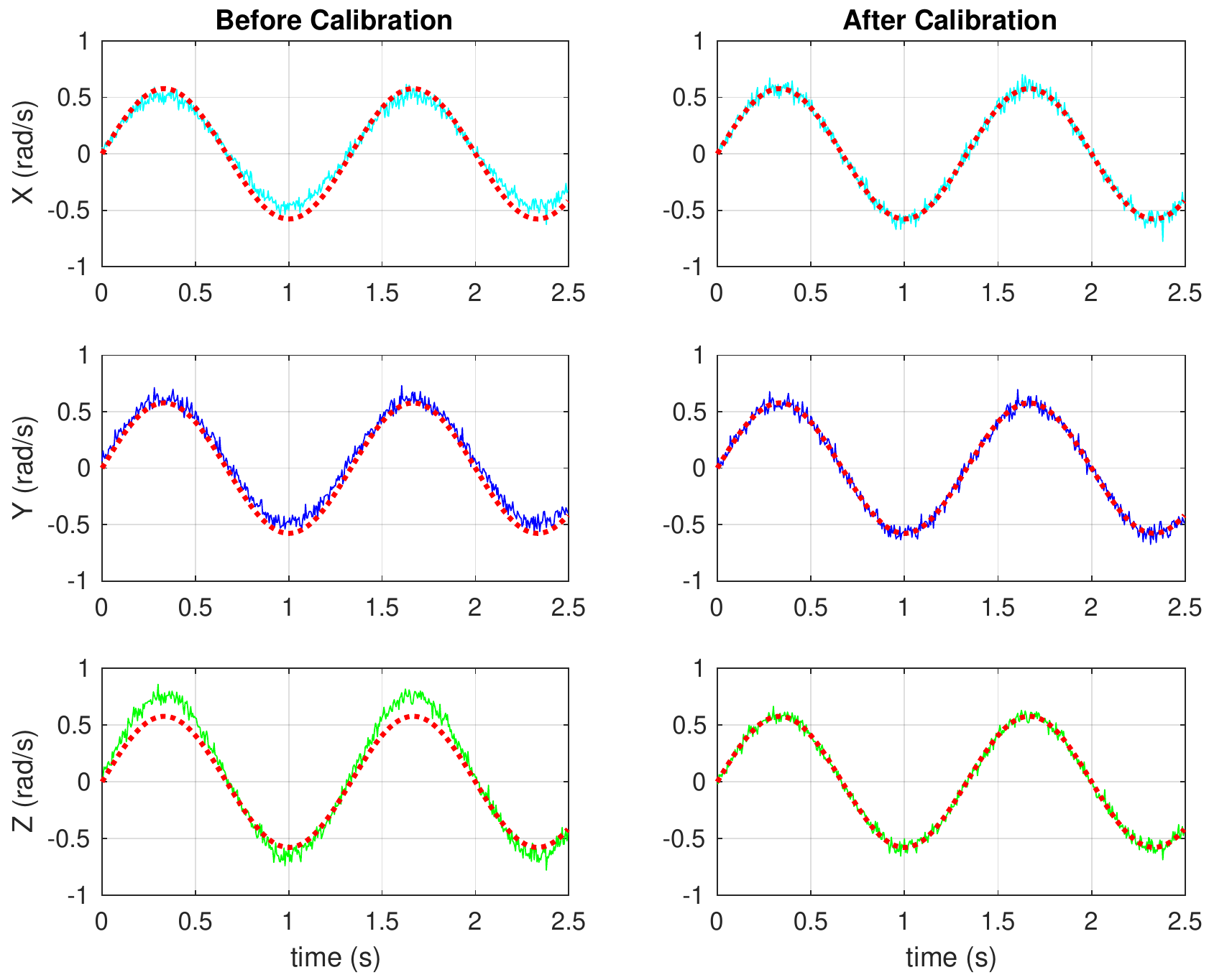}
	\caption{{Simulation results of the desired rotation speed $\omega$ and gyroscope readings from three axes $x,y,z$ before and after calibration. The dashed line indicates actual rotation on each axis, and the solid line represents gyroscope readings.}}
	\label{simrotation}
\end{figure}
We first conducted simulation tests under normal conditions. The simulation conditions are given based on the parameters of the commonly used gyroscopes.
The following assumptions on the parameter are given, and the results are explained after them.
\begin{enumerate}
  \item The scale factor follows a uniform distribution $U(80\%,120\%)$ and the bias follows $U(-0.1 rad/s, 0.1 rad/s$). The typical scale factors and biases of low-cost MEMS gyroscopes are usually within $\pm20\%$ and $\pm0.1 rad/s$, respectively.
  \item Misalignment on mounting follows $U(0\%,10\%)$. In practice, without an accurate mounting platform, it is difficult for users to make measurements in the exact position specified by the experimental protocol. To demonstrate the robustness of the proposed method, we run the simulations with mounting misalignment.
  \item The measurement noise is assumed to follow a Gaussian distribution with zero mean and two different variances, $\epsilon_1 \sim \mathcal{N}(0,\,0.035)$ and $\epsilon_2 \sim \mathcal{N}(0,\,0.2)$. The typical noise spectral density of the MEMS gyroscope is between $1.74\sim6.11\times10^{-4} rad/s/\sqrt{Hz}$. As this study uses a 200 Hz sampling rate, the range of noise amplitude is around $0.035\sim0.18 rad/s$. Thus, we consider the noise vibration as 0.035 and 0.2 $rad/s$.
  \item The variance of rotation noise is 5\% of the current speed, which follows $\mathcal{N}(0,\,5\%\omega)$. We use this term to simulate vibration during operation.
\end{enumerate}

Based on the assumptions, we generated 30 sets of scale factors and biases to simulate different gyroscopes. For each set of parameters, we repeated the simulations 500 times. For each simulation, we took a six-observation measurement according to the experimental protocol shown in Fig.\ref{rotationmethod}, and constructed a 6-by-6 observation matrix based on Eq.\eqref{observationMatrix}. Subsequently, Algorithm \ref{algorithm} was implemented to estimate the scale factors and biases. Overall, 15,000 simulations were generated for testing our proposed calibration method.

\begin{figure}
	\centering
		\includegraphics[width=0.45\textwidth]{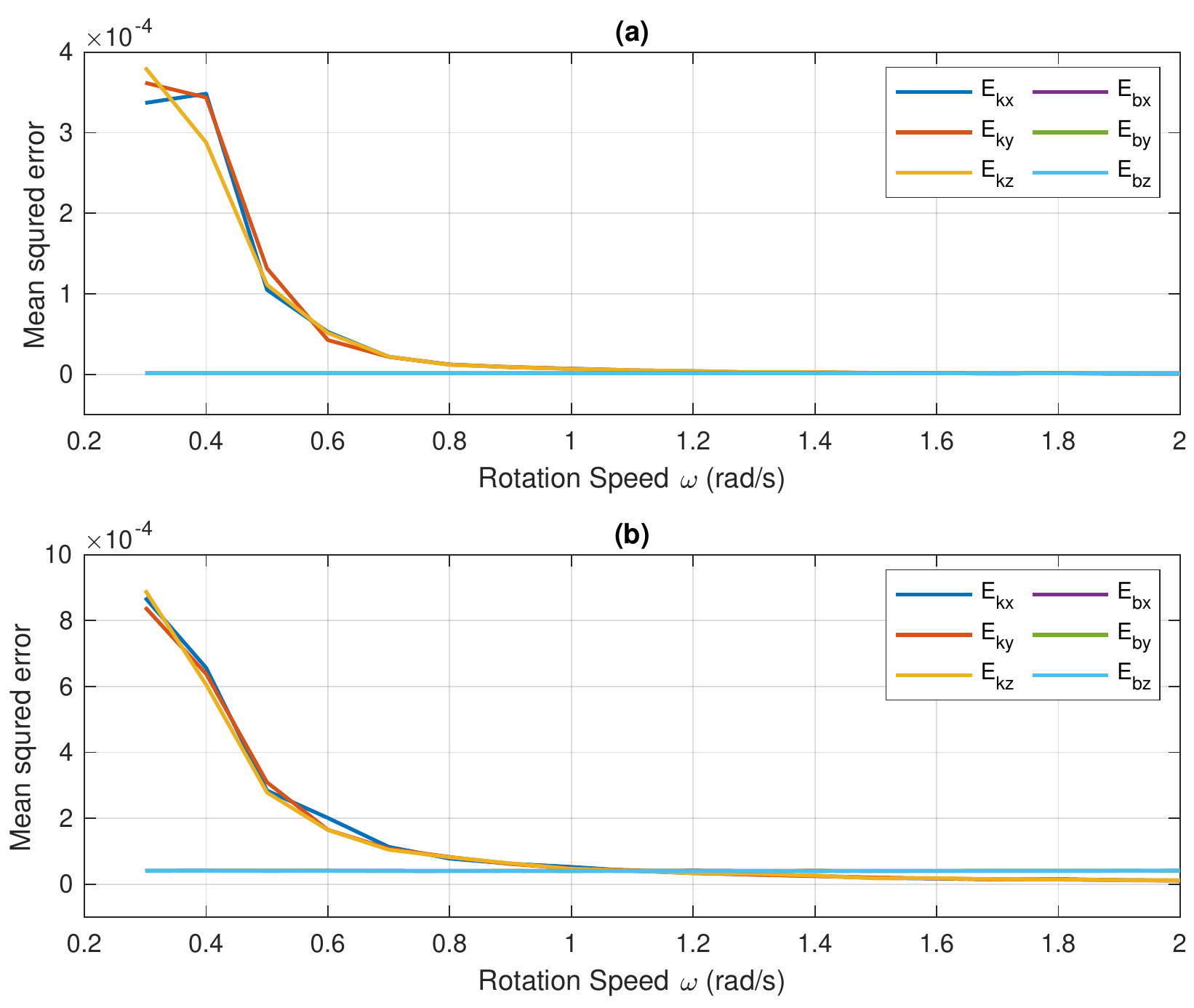}
	\caption{{The mean squared error (MSE) between estimated and actual parameters at different rotation speeds during calibration with different measurement noise levels. (a) 0.035 rad/s noise level. (b) 0.2 rad/s noise level.}}
	\label{MSE}
\end{figure}

To evaluate the performance of the proposed calibration method, we calculated the differences between the actual parameters and the estimated scale factors and biases. Box plots were used to analyze the differences as shown in Fig.\ref{boxplot}. The median values of the estimation error were 0, and the results indicated that the estimated parameters were unbiased. {The majority of estimations of scale factors had an error within $\pm9.3\times10^{-4}$ for the $0.035 rad/s$ noise level and $\pm5.4\times10^{-3}$ for the $0.2 rad/s$ noise level. The estimation error of biases is higher, which is $\pm3.0\times10^{-3}$ for the $0.035 rad/s$ noise level and $\pm1.7\times10^{-2}$ for the $0.2 rad/s$ noise level.} This indicated that the estimation accuracy was related to the measurement noise level. Better gyroscopes with lower noise have lower estimation error. {In addition, we compared Algorithm 1 with the Levenberg-Marquardt method. The results indicated that the error of these two methods were identical.} Interestingly, the scale factors had a much lower estimation error than the bias terms. This phenomenon can be explained by sensitivity analysis techniques \cite{sensitivityanalysis}. In this particular model Eq.\eqref{linearregression} and experiment design, the observability of scale factors is much higher than that of bias terms, which leads to better estimation results for the former.

To intuitively demonstrate the effectiveness of the calibration, we performed a simulation to compare the gyroscope readings before and after calibration. The desired rotation speed of the servomotor $\omega$ with respect to time was set as a sine wave with an amplitude of $1 rad/s$ and frequency of $0.75 Hz$. The three axes of the gyroscope were mounted to be equidistant from the rotation axis. Thus, the projection of the rotation speed to each axis was equal. The rotation noise variance was set to be $5\%$ of the current speed, and the measurement noise followed $\epsilon \sim \mathcal{N}(0,\,0.035)$. We randomly generated a set of parameters and use the proposed approach to estimate the scale factors and bias terms. The actual parameters $[k_x,k_y,k_z,b_x,b_y,b_z]$ and estimated parameters $[\hat{k_x},\hat{k_y},\hat{k_z},\hat{b_x},\hat{b_y},\hat{b_z}$] are as follows:
\begin{flalign*}
      \left\{
        \begin{array}{l}
            k_x=0.9070\\
            k_y=1.0501\\
            k_z=0.8734\\
            b_x=0.0528\\
            b_y=0.0813\\
            b_z=-0.0992
        \end{array}
    \right. &       \left\{
        \begin{array}{l}
            \hat{k_x}=0.9070\\
            \hat{k_y}=1.0502\\
            \hat{k_z}=0.8735\\
            \hat{b_x}=0.0529\\
            \hat{b_y}=0.0802\\
            \hat{b_z}=-0.0994
        \end{array}
    \right.
\end{flalign*}

Based on the estimated parameters, we corrected the gyroscope readings using Eq.\eqref{model}. Fig.\ref{simrotation} demonstrates that after calibration, the measured and actual values showed better fit. The fluctuations were caused by measurement noise and motor speed instability.
\begin{figure}
	\centering
		\includegraphics[scale=.5]{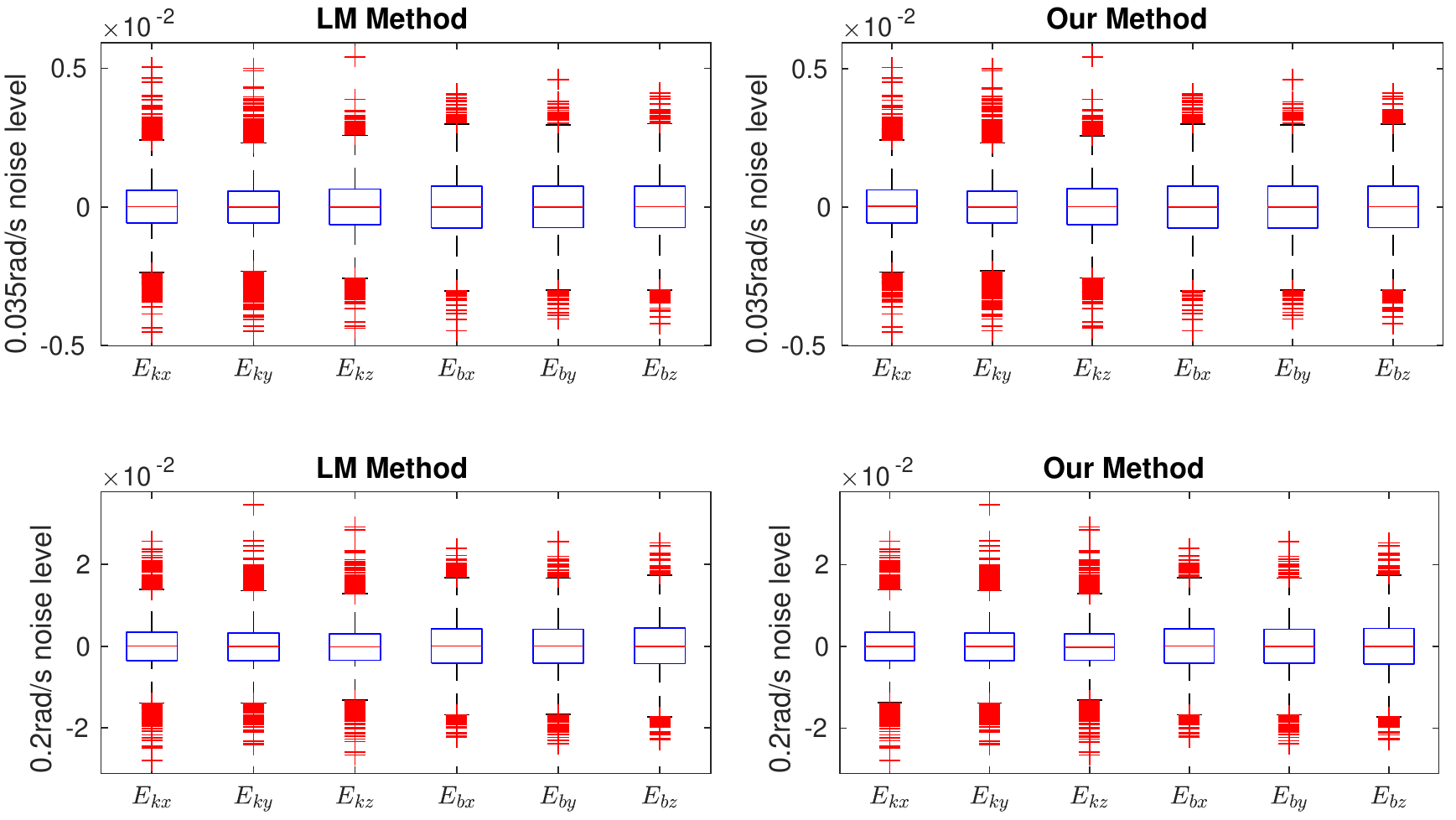}
	\caption{{The simulation results of estimation error between estimated and actual parameters under extreme conditions at different noise level using different method. Top: 0.035 rad/s noise levels. Bottom: 0.2 rad/s noise level. Left: Levenberg–Marquardt (LM) method. Right: Our proposed method.}}
	\label{boxplotextreme}
\end{figure}
\subsection{The Effect of Rotation Speed}
Next, we explored the influence of rotation speed on the estimation results during calibration. We followed the 15,000 simulations procedure described above, but used different rotation speeds. The rotation speed was set within the range of $0.3$ to $3 rad/s$ with a step size of $0.1 rad/s$. Lower rotation speeds outside this range may be covered by noise,while higher rotation speeds cannot be accurately achieved by servomotors during the real experiment. The overall mean squared error (MSE) of six estimated parameters are defined as follows:
\begin{equation}\label{sumoferror}
   e_j=\frac{1}{N}\sum_{i=1}^{N}(j_i-\hat{j_i})^2, j_i=k_{x,i},k_{y,i},k_{z,i},b_{x,i},b_{y,i},b_{z,i},
\end{equation}
where $N$ is the number of simulations. Fig.\ref{MSE} shows the influence of speed on the parameter estimation accuracy during calibration. As the speed increases, the average MSE decreases exponentially. After the speed rises to 1 rad/s, the average MSE value stops decreasing. The MSE of biases remains unchanged irrespective of changes in speed, and is only affected by the measurement noise level. This is because the observability of biases still exists even in a static state, i.e. $\omega=0$. AAppealingly, when the speed is less than 1 rad/s, the MSE of scale factors drops significantly as the speed increases. At low speeds, the measurement noise occupies most of the measured value rather than the projection of the rotation component on this axis. At this time, the signal to noise ratio (SNR) of the measured value is small. The lower the rotation speed, the smaller the SNR. At the same rotation speed, when comparing (a) and  (b) in Fig.\ref{MSE}, the estimation with high measurement noise has a larger MSE.
\begin{figure}[t]
	\centering
		\includegraphics[width=0.45\textwidth]{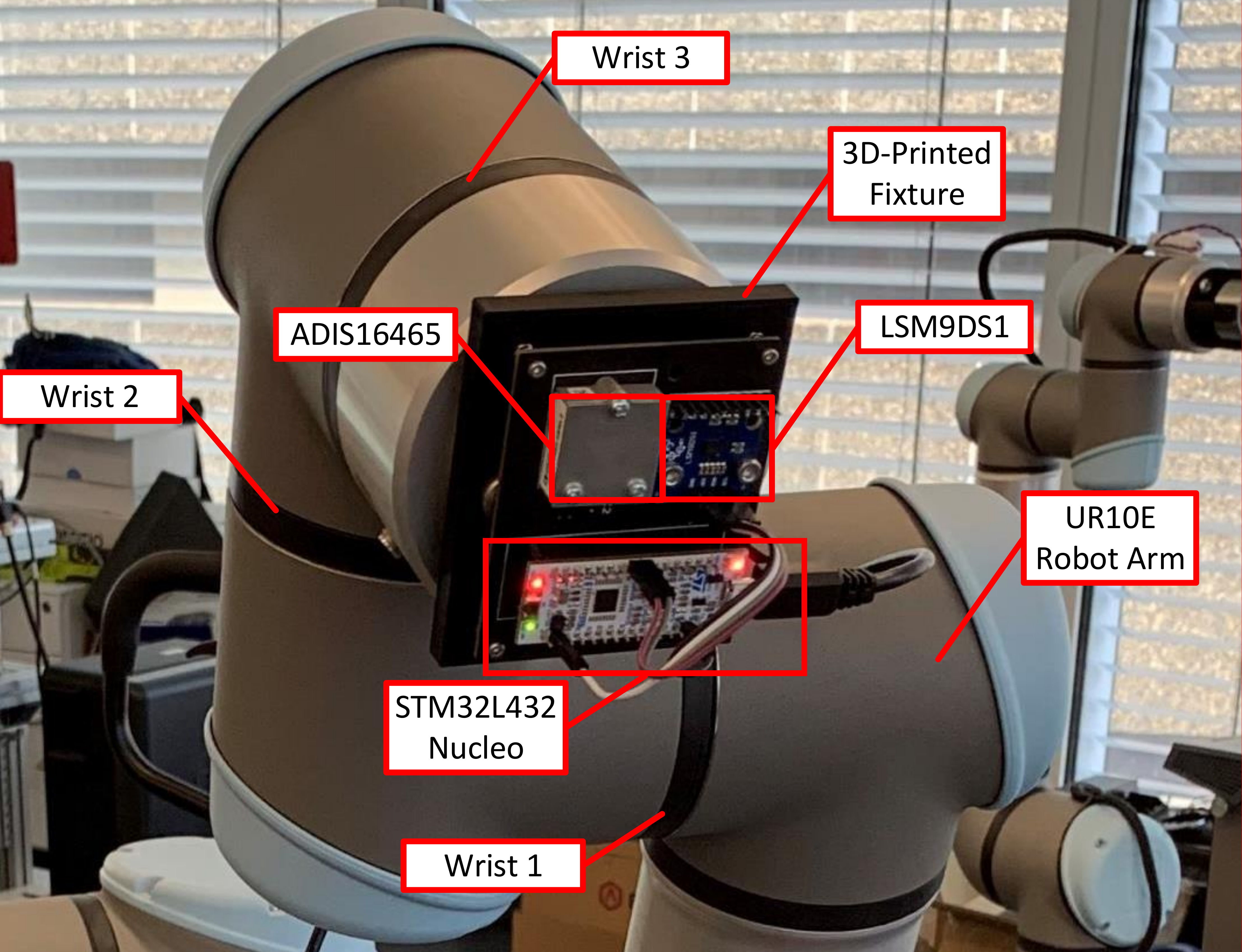}
	\caption{Experimental system for the gyroscope calibration on a robot arm UR10e. The part names and joint numbers are noted.}
	\label{system}
\end{figure}

\subsection{Robustness of the Method under Extreme Conditions}
\begin{table}[h]
  \centering
  \caption{Convergence rate under different scale factors and biases}\label{convergence}
  \resizebox{0.48\textwidth}{!}{
    \begin{tabular}{ccccccc}
      \toprule
      Number of\\ iterations & $k_x$ & $k_y$ & $k_z$ & $b_x$ & $b_y$ & $b_z$ \\
      \hline
       & \multicolumn{6}{c}{1.High scale factor error}\\ 
      Actual value & 1.9074 & 1.9529 & 1.5635 & 0.0827 & 0.0265 & -0.0805 \\
      1& 1.9102&	1.9571&	1.5665&	0.0822&	0.0243&	-0.0797\\
      2-Converged & 1.9071 &   1.9539 &   1.5640 &   0.0822   & 0.0243 &  -0.0797 \\
      \hline
      & \multicolumn{6}{c}{2.High bias}\\
      Actual value &   1.0979&    1.1052 &   0.9851 &  -0.1046 &   0.1995 &   0.1565 \\
      1          & 1.1029   & 1.1103  &  0.9900  & -0.1057  &  0.1971  &  0.1545\\
      2          & 1.0961    &1.1035 &   0.9839 &  -0.1057 &   0.1971 &   0.1545 \\
      3-Converged          & 1.0962    &1.1036&    0.9840&   -0.1057&    0.1971&    0.1545 \\
      \hline
      & \multicolumn{6}{c}{3.High scale factor error and bias}\\
     Actual value &  1.5044    &1.6494   & 1.5282 &   0.1483  & -0.1282&    0.1794\\
      1 & 1.5173  &  1.6652   & 1.5423&    0.1469  & -0.1284&    0.1802 \\
      2 &  1.5053  &  1.6521 &   1.5302&    0.1469&   -0.1285&    0.1803\\
      3-Converged & 1.5055    &1.6523 &   1.5304  &  0.1469&   -0.1285  &  0.1803 \\
      \hline \\
    \end{tabular}}
\end{table}
To demonstrate the robustness of the proposed gyroscope calibration method, the quality of the gyroscope was assumed to be very poor. The randomly generated parameters followed $U(120\%,200\%)$ for scale factors and $U(-0.2\sim -0.1,0.1\sim0.2)$ for biases. Other parameters followed the previous setting. The results shown in Fig.\ref{boxplotextreme} suggested that our proposed method could be applied to gyroscopes with poor manufacturing quality. Although the errors were larger than under normal conditions, the majority of these errors were within $\pm3\times10^{-3}$ for 0.035 rad/s noise level and $\pm1.8\times10^{-2}$ for 0.2 rad/s noise level. Under extreme conditions, the scale factors had worse observability since larger scale factors enlarge the signal noise, thereby reducing the SNR.

\begin{figure}
	\centering
		\includegraphics[width=0.45\textwidth]{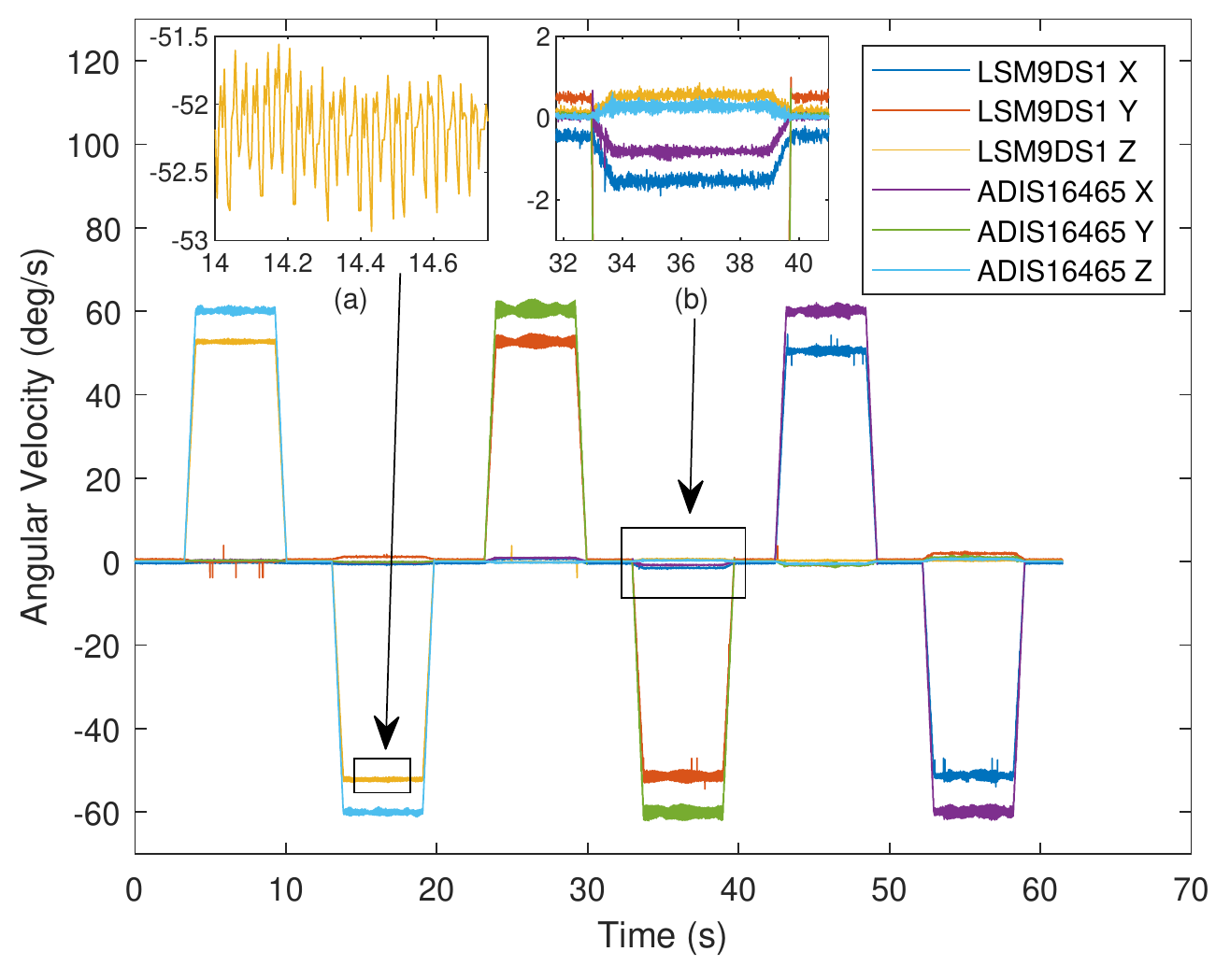}
	\caption{Raw gyroscope data from LSM9DS1, compared with ADIS16465 reading during calibration. (a) Periodic vibration was caused by the control strategy of the servomotor. (b) The component on the non-rotating axis was caused by mounting misalignment.}
	\label{rawdataL9}
\end{figure}
\begin{figure}
	\centering
		\includegraphics[width=0.45\textwidth]{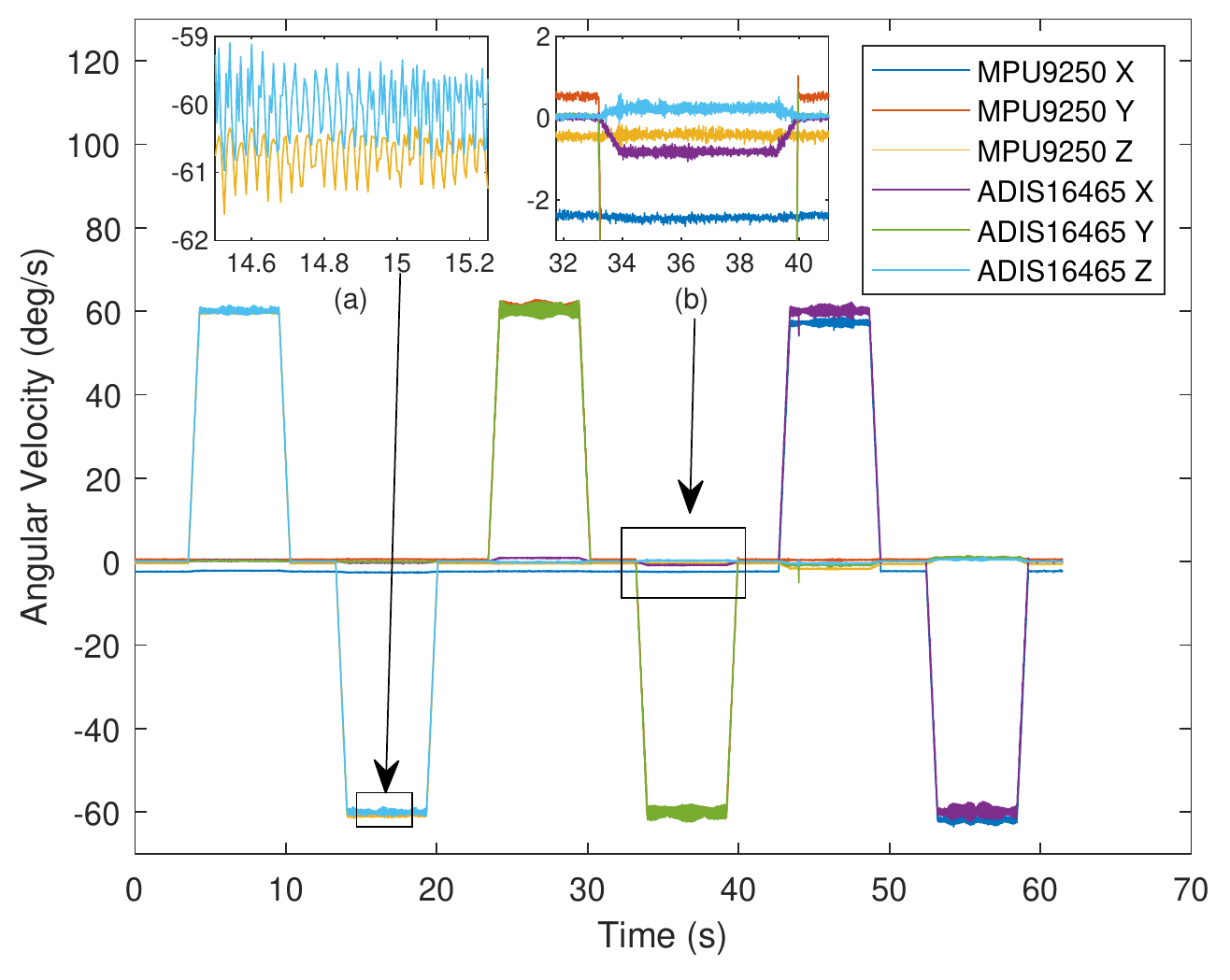}
	\caption{Raw gyroscope data from MPU9250, compared with ADIS16465 reading during calibration. (a) Periodic vibration was caused by the control strategy of the servomotor. (b) The component on the non-rotating axis was caused by mounting misalignment.}
	\label{rawdataM9}
\end{figure}

To demonstrate the convergence rate of the iterative method, we performed three simulations (results shown in TABLE \ref{convergence}). The first simulation used high scale factors error and normal biases, while the second simulation used typical scale factors error and high biases. The third simulation used high scales factors error and high biases. The results indicated that less than three iterations were needed for the proposed calibration method.

\section{Experiment} \label{Sec:Experiment}
\begin{figure}
	\centering
		\includegraphics[width=0.45\textwidth]{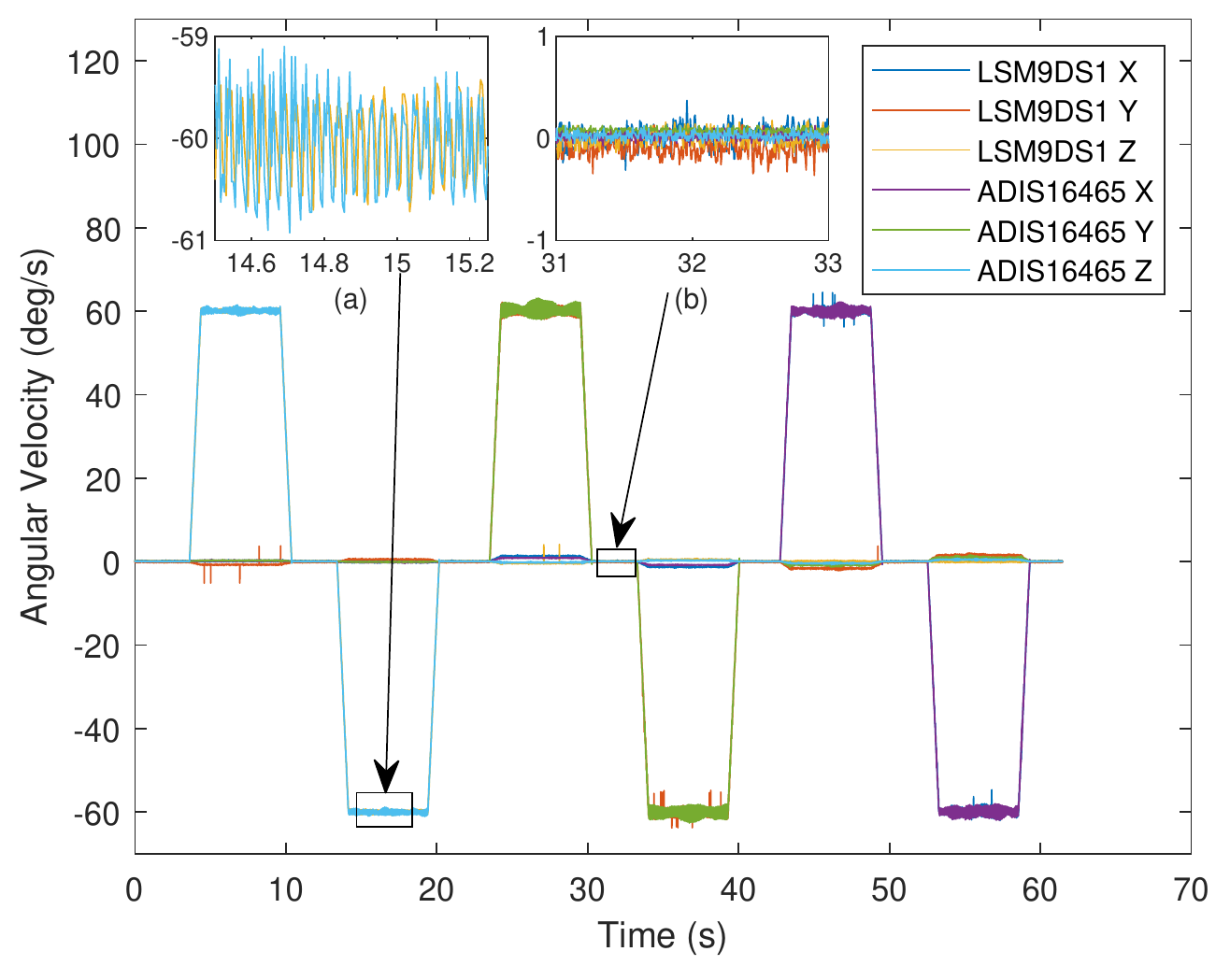}
	\caption{Calibrated gyroscope data from LSM9DS1, compared with ADIS16465 reading during the testing period. (a) The reading from LSM9DS1 and ADIS16465 nearly coincided with each other. (b) The biases of gyroscope reading were almost zero.}
	\label{caldataL9}
\end{figure}
\begin{figure}
	\centering
		\includegraphics[width=0.45\textwidth]{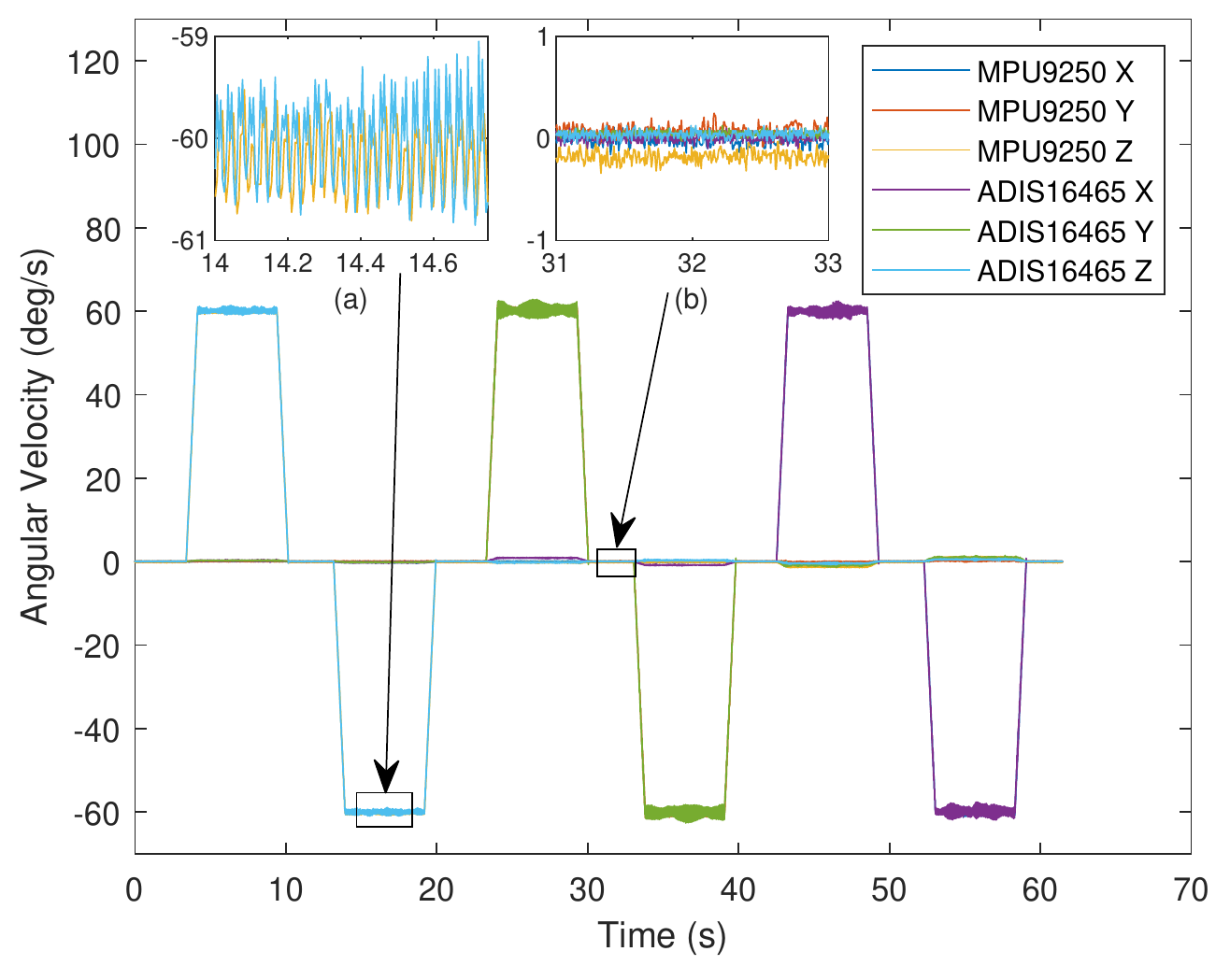}
	\caption{Calibrated gyroscope data from MPU9250, compared with ADIS16465 reading during the testing period. (a) The reading from MPU9250 and ADIS16465 nearly coincided with each other. (b) The biases of gyroscope reading were almost zero.}
	\label{caldataM9}
\end{figure}

We calibrated and verified two commonly used low-cost MEMS gyroscopes: LSM9DS1 from STMicroelectronics and MPU9250 from TDK. We demonstrated the application of our proposed method on a UR10e robotic arm. {Besides, our method is easy to implement on a 3-axis camera gimbal or a single-axis servomotor with proper adapter.} The calibration system is shown in Fig.\ref{system}. The gyroscope is replaceable. The digital signals were collected and calculated using an STM32L432 Nucleo board. It is worth noting that high-precision turntables or other calibration equipment were not needed in the proposed system. For the purpose of comparing the data quality of low-cost gyroscopes, an ADIS16465 sensor was mounted on the same board during calibration. The ADIS16465 was pre-calibrated using a turntable. The room temperature was set to 22 \textdegree C.

\subsection{Calibration of Two Low-cost Gyroscopes}

The proposed method was applied to the two gyroscopes according to the G-optimal experiment scheme shown in Fig.\ref{rotationmethod}. We summarise the calibration procedure as follows:
\begin{enumerate}
    \item Mount the LSM9DS1 and ADIS16465 on the UR10e and turn on the system.
    \item Rotate wrist 3 of the UR10e 360$^{\circ}$ clockwise. Based on the simulation results, find the balance between the rotation speed  and measurement noise, and set the angular velocity to 60$^{\circ}$/s. Then, wait for 3 seconds.
    \item Rotate wrist 3 of the UR10e 360$^{\circ}$ counterclockwise. Wait for 3 seconds. 
    \item Repeat 2-3 for wrist 1 and wrist 2.
    \item Repeat 1-4 for MPU9250.
\end{enumerate}
The entire process takes about one minute. The data was recorded by the microcontroller and transferred to the computer via a serial port. After completing the calibration process on the microcontroller, the scale factors and biases were transferred to the computer. The raw data from LSM9DS1 and MPU9250 are shown in Fig.\ref{rawdataL9} and Fig.\ref{rawdataM9}.

From the figures, we can see that the readings of the two sensors are quite different from that of ADIS16465. Since the ADIS16465 is a high-precision sensor and we have also pre-calibrated it, it can be inferred that the difference is caused by the scale factors and biases of the low-cost sensors. The servomotor control strategy leads to vibration during rotation, while the ADIS16465 reading suggests that the average angular velocity is relatively accurate. Since high-precision components were not used in the mounting process, the non-rotating axis also had a rotation component. We hence prove that the proposed method can work under conditions of vibration and misalignment.

To demonstrate the accuracy of the proposed method, we compared the calibration results with those obtained using the precision turntable method \cite{conventionalmethod}. The results are shown in TABLE \ref{resultsL9} and TABLE \ref{resultsM9}.
\begin{table}[h]
  \centering
  \caption{Comparison of LSM9DS1 calibration results}\label{resultsL9}
    \begin{tabular}{ccc}
      \toprule
      Parameter & Results of proposed   & Results of conventional  \\
      & calibration method & turntable method\\
      \hline
      $k_x$ & 1.1775 & 1.1771  \\
      $k_y$& 1.1552&	1.1554\\
      $k_z$ & 1.1445 &   1.1440  \\
      $b_x(rad/s)$ & 0.0076 & 0.0077  \\
      $b_y(rad/s)$& -0.0103&	-0.0100\\
      $b_z(rad/s)$ & -0.0042 &   -0.0051  \\
      \hline \\
    \end{tabular}
\end{table}
\begin{table}[h]
  \centering
  \caption{Comparison of MPU9250 calibration results}\label{resultsM9}
    \begin{tabular}{ccc}
      \toprule

      Parameter & Results of proposed   & Results of conventional  \\
      & calibration method & turntable method\\
      \hline
      $k_x$ & 1.0069 & 1.0065  \\
      $k_y$& 0.9960&	0.9955\\
      $k_z$ & 0.9955 &  0.9950  \\
      $b_x(rad/s)$ & 0.0442 & 0.0411  \\
      $b_y(rad/s)$& -0.0089&	-0.0099\\
      $b_z(rad/s)$ & 0.0076 &   0.0101  \\
      \hline \\
    \end{tabular}
\end{table}
For a fast calibration method without using any high-precision equipment, all errors less than $10^{-3}$ indicate a considerably accurate result. Besides, considering the low repeatability of the two gyroscopes used, the actual scale factors and biases when mounted on the robot arm may be different from those when mounted on the turntable. 
\begin{table}
  \centering
  \caption{MSE between LSM9DS1 and ADIS16465}\label{RMSL9}
    \begin{tabular}{ccc}
      \toprule

      Axis & Error before   & Error after \\
      & calibration (rad/s) & calibration (rad/s)\\
      \hline
      $x$ & 0.2798 & 0.0052  \\
      $y$& 0.2255&	0.0093\\
      $z$ & 0.1940 &  0.0022  \\
      \hline \\
    \end{tabular}
\end{table}
\begin{table}
  \centering
  \caption{MSE between MPU9250 and ADIS16465}\label{RMSM9}
    \begin{tabular}{ccc}
      \toprule
      Axis & Error before   & Error after \\
      & calibration (rad/s) & calibration (rad/s)\\
      \hline
      $x$ & 0.1031 & 0.0051  \\
      $y$& 0.0107&	0.0057\\
      $z$ & 0.0096 &  0.0033  \\
      \hline \\
    \end{tabular}
\end{table}

After calibration, the parameters were stored in the microcontroller. To further test calibration effectiveness, we made the robot arm repeat the same movements as during the calibration procedure. Instead of raw data, the microcontroller sent the calibrated gyroscope readings to the computer. The results after calibration are shown in Fig.\ref{caldataL9} and Fig.\ref{caldataM9}. The reading from the pre-calibrated ADIS16465 was considered as a ground truth. We calculated the MSE error between the LSM9DS1/MPU9250 and ADIS16465. The results are shown in TABLE \ref{RMSL9} and TABLE \ref{RMSM9}. Our experiments show that the gyroscope reading obtained after calibration using the proposed method is significantly more accurate than the reading before calibration.

\section{Conclusion} \label{Sec:Conclusion}
This paper proposed an efficient servomotor-aided calibration method that estimates the gain factors and biases of a triaxial gyroscope. A six-observation G-optimal experimental scheme was implemented for the calibration process, and a fast converging recursive linear least square estimation method was applied to reduce the computational complexity. We performed a series of simulations and experiments to evaluate the validity and feasiblity of the proposed method.

The simulation results indicated that the gyroscope parameters could be accurately estimated within three iterations, and demonstrated that the proposed method was robust under extreme conditions. Furthermore, the simulation results showed a balance between the rotation speed and measurement noise. The angular velocity in the experiment was set to 60$^{\circ}$/s, accordingly. 

We performed experiments on two commonly used low-cost MEMS gyroscopes. The outcomes of calibration using our proposed method and the conventional turntable method were experimentally compared. The results indicated that the errors between these two methods were less than $10^{-3}$. To further test the performance of our proposed method, we compared the calibrated reading of the two low-cost gyroscopes with a high-precision sensor. The results showed that the error was significantly decreased after calibration. More importantly, we demonstrated the possibility of implementing this method on low-precision motors such as a robot arm, as well as its applicability on a microprocessor. Using our proposed method, the entire calibration process only requires one minute, and high-precision calibration equipment is not necessary.

\bibliographystyle{IEEEtran}

\bibliography{IEEEexample}

\begin{IEEEbiography}[{\includegraphics[width=1in,height=1.25in,clip,keepaspectratio]{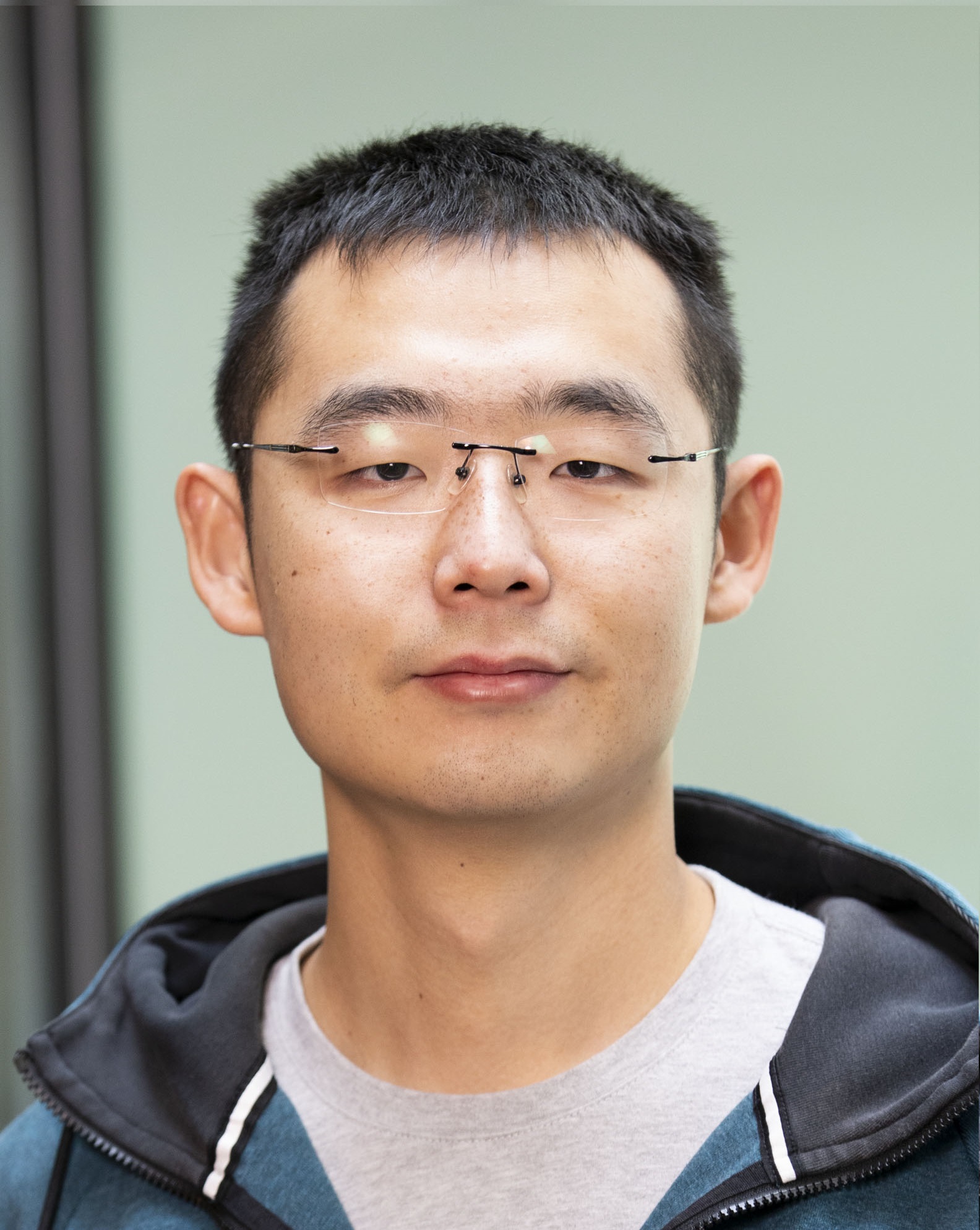}}]{Li Wang}
received the Bachelor of Engineering degree from the Wuhan University in 2016 and Master of Engineering degree from the Australia National University in 2019. He is currently a PhD student at Faculty of Engineering and Information Technology, UTS, and perusing his research career in Biomedical Engineering. Li’s research interests are on embedded system design, wearable health monitoring system design, IMU calibration, cardiovascular rehabilitation system design and reinforcement learning control.
\end{IEEEbiography}
\vspace{-2cm}

\begin{IEEEbiography}[{\includegraphics[width=1in,height=1.25in,clip,keepaspectratio]{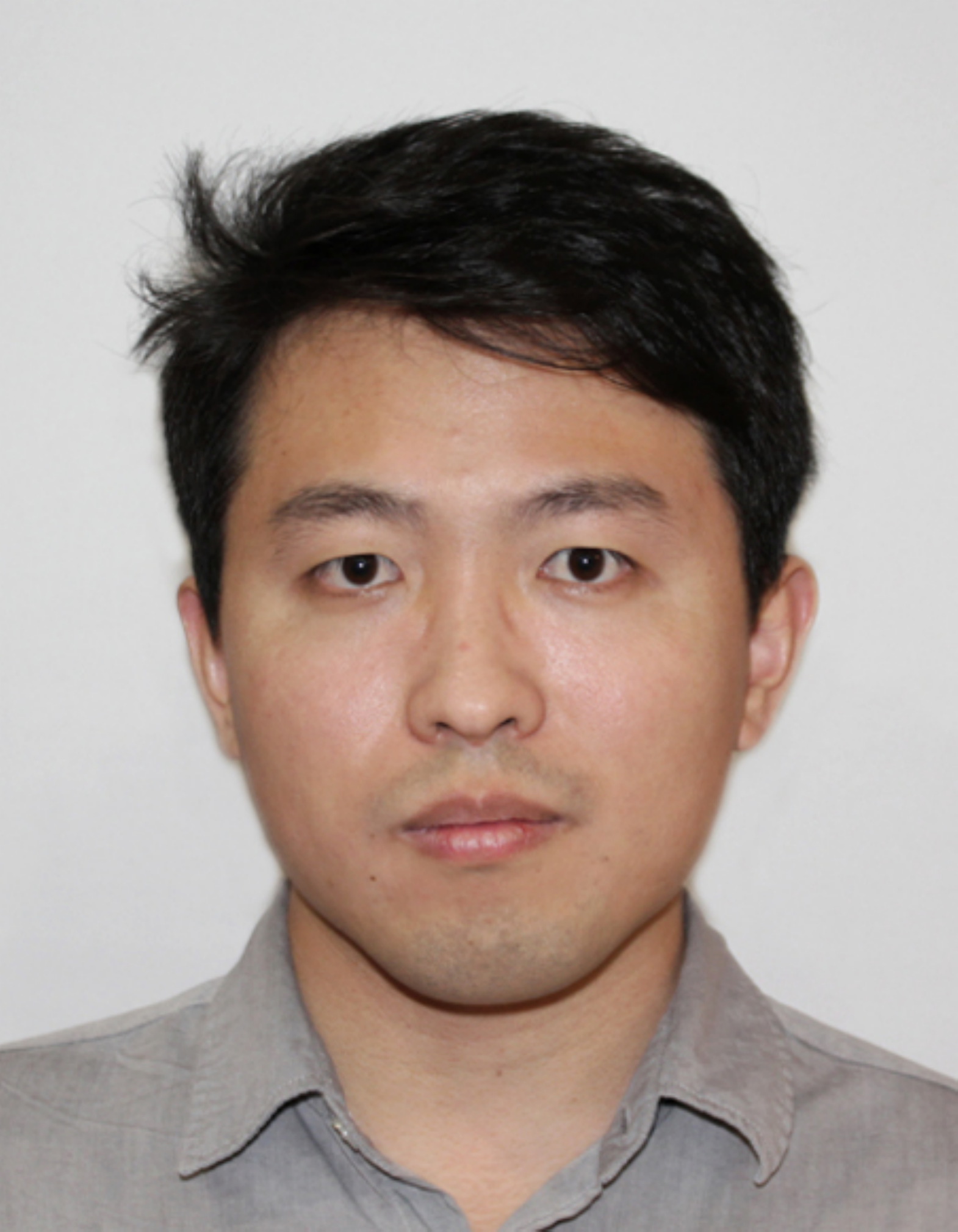}}]{Tao Zhang}(S'13$-$M'20) received the B.S. and M.S. degrees in mechanical engineering and mechatronics engineering from the Huazhong Agricultural University, Wuhan, China, and the Harbin Institute of Technology, Shenzhen, China, in 2009 and 2012, respectively, and the Ph.D. degree from the Faculty of Engineering and Information Technology, University of Technology Sydney, NSW, Australia, in 2019. 
\end{IEEEbiography}
\vspace{-2cm}

\begin{IEEEbiography}[{\includegraphics[width=1in,height=1.25in,clip,keepaspectratio]{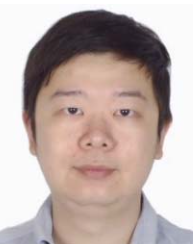}}]{Lin Ye} received the B.E. (Hons. I) degree in electrical engineering from the University of Technology, Sydney, Australia, in 2013, and the Ph.D. degree from the Center for Health Technologies, University of Technology, in 2018. He worked at Intelligent Driving Development Center, Geely Automobile Research Institute Co.,Ltd until 2020.  Currently, he works at Enjoymove Technology Co., Ltd as senior algorithm engineer. His current research interests include system identification, path planning and control of autonomous vehicle.
\end{IEEEbiography}
\vspace{-2cm}

\begin{IEEEbiography}[{\includegraphics[width=1in,height=1.25in,clip,keepaspectratio]{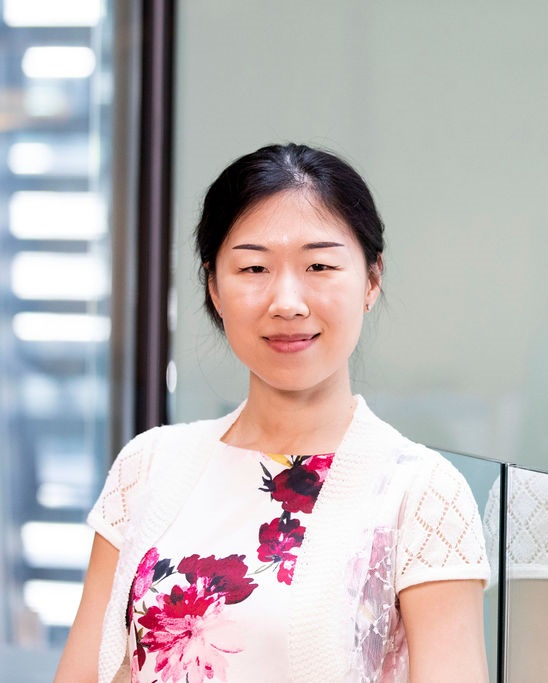}}]{Jiao Jiao Li} received her Bachelor of Biomedical Engineering (Honours I with University Medal)/Bachelor of Medical Science and PhD from the University of Sydney in 2010 and 2015, respectively. She is currently a Lecturer in Biomedical Engineering at UTS, a National Health and Medical Research Council (NHMRC) Early Career Fellow, a chief investigator on the Australian Research Council Training Centre for Innovative BioEngineering, and a Science \& Technology Australia 2021-22 Superstar of STEM. Her research in regenerative medicine focuses on using stem cells and other approaches to treat chronic musculoskeletal conditions such as osteoarthritis.
\end{IEEEbiography}
\vspace{-2cm}

\begin{IEEEbiography}[{\includegraphics[width=1in,height=1.25in,clip,keepaspectratio]{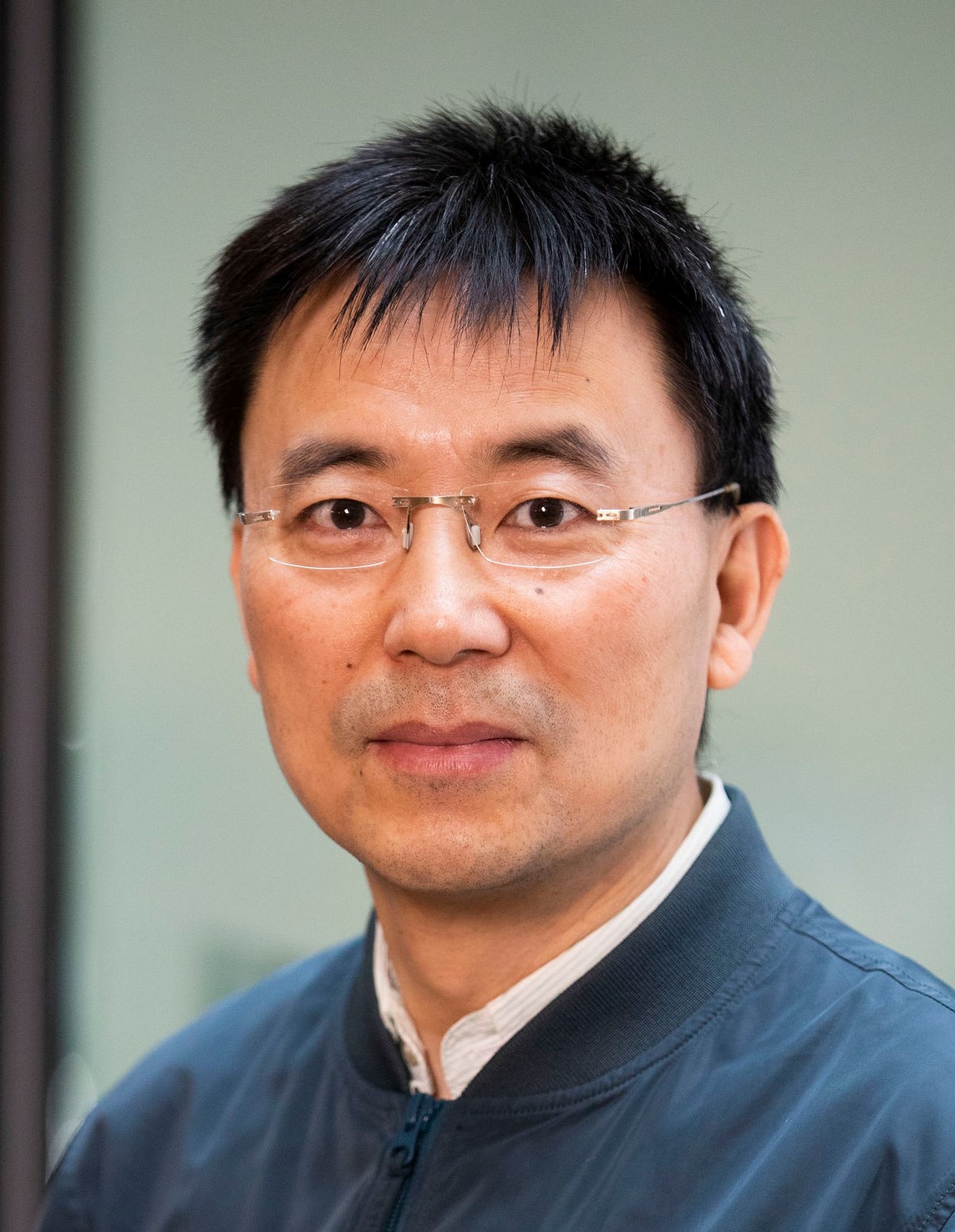}}]{Steven W. Su} (M'99-SM'17) received the B.S. and M.S. degrees from the Harbin Institute of Technology, Harbin, China, in 1990 and 1993, respectively, and the Ph.D. degree from the Research School of Information Sciences and Engineering, Australian National University, Canberra, Australia, in 2002.

He was a Post-Doctoral Research Fellow with the Faculty of Engineering, University of New South Wales, Sydney, NSW, Australia, from 2002 to 2006. He is currently an Associate Professor with the Faculty of Engineering and Information Technology, University of Technology Sydney, Sydney, Australia. His current research interests include biomedical system modeling and control, nonlinear robust control, fault tolerant control, and wearable monitoring systems.

\end{IEEEbiography}

\end{document}